\documentclass[conference]{IEEEtran}
\IEEEoverridecommandlockouts

\usepackage{color}
\usepackage{graphicx}
\usepackage{amsmath}
	
\usepackage{soul}

\usepackage{subcaption}
\usepackage{caption}
\usepackage[utf8]{inputenc}

\makeatletter
\def\hlinewd#1{%
\noalign{\ifnum0=`}\fi\hrule \@height #1 %
\futurelet\reserved@a\@xhline}
\makeatother
\usepackage{booktabs}
\usepackage{multirow}

\usepackage[binary-units=true]{siunitx}

\usepackage[algoruled,noline,longend,linesnumbered]{algorithm2e}

\usepackage{float}

\usepackage{pgfplots}
\usepackage{pgfplotstable}
\usepackage{siunitx}

\makeatletter

\usepackage{amsthm}

\usepackage[small,compact]{titlesec} 
\titlespacing{\section}{2pt}{2pt}{2pt}
\titlespacing{\subsection}{2pt}{2pt}{2pt}
\titlespacing{\subsubsection}{2pt}{2pt}{2pt}

\begin{document}

\graphicspath{{Fig/}}
\def\figname{Figure}
\def\algname{Algorithm}

\newcommand{\papertitle}{Early-Exit with Class Exclusion for Efficient Inference of Neural Networks
}

\title{\papertitle}

\author{
\IEEEauthorblockN{Jingcun Wang$^1$,
Bing Li$^2$,
Grace Li Zhang$^1$}
\IEEEauthorblockA{$^1$Hardware for AI Group, TU Darmstadt, Germany, $^2$Chair of Electronic Design Automation, TU Munich, Germany}
\IEEEauthorblockA{Email: \{jingcun.wang, grace.zhang\}@tu-darmstadt.de, b.li@tum.de}
}

\maketitle



\begin{abstract}

Deep neural networks (DNNs) have been successfully applied in various fields. In DNNs, a large number of multiply-accumulate (MAC) operations are required to be performed, posing critical challenges in applying them in resource-constrained platforms, e.g., edge devices. 
To address this challenge, in this paper, we propose a class-based early-exit for dynamic inference. Instead of pushing DNNs to make a dynamic decision at intermediate layers, we take advantage of the learned features in these layers to exclude as many irrelevant classes as possible, %
so that later layers only have to determine the target class among the remaining classes. 
When only one  class remains at a layer, this class is the corresponding classification result. 
Experimental results demonstrate the computational cost of DNNs in inference can be reduced significantly with the proposed early-exit technique. 
The codes can be found at https://github.com/HWAI-TUDa/EarlyClassExclusion.

\end{abstract}

\section{Introduction} \label{sec:intro} 
In the past decade, deep neural networks (DNNs) have achieved remarkable breakthroughs in various fields, e.g., image classification
and object detection. In DNNs, a large number of floating-point operations (FLOPs), mainly consisting of multiply–accumulate (MAC) operations, need to be executed. 
For example, ResNet50 \cite{resnet} requires 4.1G MAC operations and around 8.2G FLOPs to predict a classification result for a $224\times224$ image. This tremendous computational cost 
poses critical challenges in applying DNNs in resource-constrained hardware platforms, e.g., edge devices.

To reduce the computational cost of executing DNNs, 
various techniques have been introduced at software level. For example, 
pruning \cite{pruninclass}\cite{Petri2023PowerPruningSW} reduces the number of MAC operations and FLOPs by removing  unnecessary weights. Quantization \cite{qinference}\cite{sun2023class} approximates floating-point MAC operations with fixed-point ones to reduce the computational complexity of MAC operations.
Knowledge distillation \cite{distilling}\cite{ruidi24} transfers a large DNN model to a compact model consisting of few MAC operations. Efficient DNN architectures such as MobileNet \cite{depth_conv} 
and SteppingNet \cite{10136943} 
have also been introduced to improve the computational efficiency. 
Neural architecture search \cite{nas} explores more efficient neural network architectures. 

The previous solutions described above assume the structure of a DNN model is static, indicating that all input data need to flow to the end of the DNN. Such a static structure leads to a low computational efficiency in processing different inputs. To address this issue, previous work developed dynamic neural networks \cite{Han2021DynamicNN}, attempting to adapt their structures, e.g., depth, according to different inputs. 
Neural networks with early-exit at their intermediate layers 
are a representative type of dynamic neural networks. 
Such an early-exit dynamic network 
incorporates multiple output branches, also called exit points, at intermediate layers of the network. In inference, 
if the feature maps at an exit point are sufficient to make a correct classification, 
the corresponding classifier determines the class with the largest probability and the inference terminates to reduce computational cost. 


Previous work has explored early-exit dynamic neural networks with various techniques. For example, 
BranchyNet \cite{branchynet} manually inserts two exit points at pre-defined intermediate layers of DNNs. 
MSDNet \cite{multiscale} uses early-exit to enable 
 an anytime/budgeted classification for DNNs. \cite{li2019improved}  enhances the accuracy of  early-exit dynamic networks by applying gradient equilibrium and one-for-all knowledge distillation. 
 EPNet \cite{dai2020epnet} develops a lightweight early-exit structure and determines the exit policy using a Markov decision process. Shallow-Deep Networks \cite{shallow} exploit early-exit to address the overthinking problem in neural networks. 
\cite{han2022learning} 
proposes 
a sample weighting technique to enhance the accuracy of 
classifiers at each exit point.
EENet \cite{ilhan2023eenet} uses a multi-objective training to fine-tune an early-exit policy to enhance the computational efficiency.


In the previous early-exit work described above, once the learned features at a layer are not able to decide the correct class, all the intermediate computation results are discarded and the exit condition is evaluated at the next layer anew. Accordingly, some computation effort conducted at early layers may be wasted, which leads to a low computation efficiency. To overcome this problem, we propose a class-based early-exit strategy for dynamic inference in which the learned features in early layers 
are exploited to exclude as many irrelevant classes as possible, so that the target class can be rapidly revealed before the last layer is reached. 
An early decision can  be made when only one class remains. 
\textit{A class is defined as an image category, e.g., cat.} 
\textit{An excluded class means that it can no longer be considered as the correct class}. 

The rest of this paper is organized as follows. Section \ref{sec:PreliminariesMotivation}
describes the background and motivation of this work. Section \ref{sec:methods} explains the proposed class-based early-exit framework for dynamic inference. Section \ref{sec:results} and \ref{sec:conclusion} respectively present the experimental results and conclusions.

\section{Background and Motivation}
\label{sec:PreliminariesMotivation} 



\begin{figure}
\centering    
\includegraphics{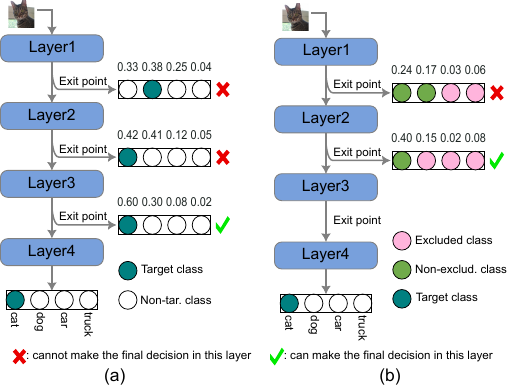}
\caption{(a) Traditional early-exit strategy. (b) The proposed class-exclusion early-exit strategy.}
\label{fig:motivation_fig}
\end{figure}

Early-exit, which allows ``easy" inputs to be classified and exit at shallow layers without executing deeper layers, 
is one strategy to realize dynamic neural networks. An early-exit dynamic neural network consists of two parts, a backbone network and multiple output branches, also called exit points, as illustrated in Fig. \ref{fig:motivation_fig}(a). 
A backbone could be any neural networks, such as AlexNet \cite{alexnet}, ResNet \cite{resnet}. An exit point typically consists of a simple classification neural network that takes compressed feature maps from the backbone as inputs and uses SoftMax function as its output activation function to generate prediction probabilities. If the feature maps at an exit point are sufficient to make a correct classification, 
the corresponding classifier determines the class with the largest probability, and the inference terminates, as shown in Fig. \ref{fig:motivation_fig}(a).

In previous early-exit techniques,  
once the learned features at a layer are not able to decide the correct class, all the intermediate computation results are discarded, leading to a waste 
of computational information. 
To overcome their drawback, we take advantage of the learned features at early layers to exclude as many irrelevant classes as possible, 
so that later layers only have to determine the target class among the remaining classes. 
Excluding irrelevant classes with the learned features at intermediate layers is viable since such features can obviously differentiate different classes. For example, the feature of wheel learned at one intermediate layer can directly determine that an input does not belong to the classes cat and dog. 
As more and more irrelevant classes are excluded, the classification decision becomes clear, facilitating a rapid early exit. Till only one class remains, the classification result is obtained naturally. 


Fig. \ref{fig:motivation_fig}(b) illustrates the concept of the class-based early-exit for dynamic inference. 
Different from traditional early-exit in Fig. \ref{fig:motivation_fig}(a), the irrelevant classes are excluded by exploiting the learned feature maps in the first two layers. Since there is only one remaining class after class exclusion at the second layer,  the remaining class is thus the target class. According to this figure, the class-based early-exit strategy has a good potential to exit earlier than the traditional early-exit techniques, leading to better computational efficiency.



\section{Class-Based Early-Exit Framework}
\label{sec:methods}
In this section, we will introduce the proposed class-based early-exit framework in detail. 


\begin{figure}
\centering    
\includegraphics{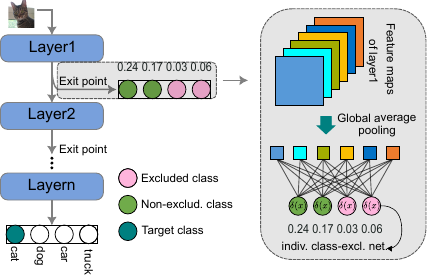}
\caption{Construction of class-exclusion neural networks for four classes.}
\label{fig:classifier}
\end{figure}  

\subsection{Construction of Class-Exclusion Neural Networks for Excluding Irrelevant Classes}

To exclude irrelevant classes at intermediate layers, we exploit the learned features in these layers to construct class-exclusion neural networks.
Since whether one class is excluded or not should not be  dependent on the other classes, we assign an individual class-exclusion network to each class to determine whether this class can be ruled out for an input, as illustrated in Fig. \ref{fig:classifier}. To reduce the computational cost incurred by the class-exclusion networks, a simple fully-connected layer is used as the structure of such a network. 

For the inputs of each class-exclusion network at an intermediate layer, the feature maps at this layer can be flattened and used as the inputs. However, the large size and the large number of feature maps at an intermediate layer lead to a high computational cost. To address this issue, the feature maps at an intermediate layer are compressed with global average pooling operations, as shown in  
Fig. \ref{fig:classifier}. 

For the output of each class-exclusion network, the following Sigmoid function in Equation~\ref{eq:sigmodi} instead of SoftMax is used as the activation function 
\begin{equation}
\label{eq:sigmodi}
\delta(x)=\frac{1}{1+e^{-x}} 
\end{equation}
where $x$ is the output of the corresponding neuron in the class-exclusion network. 
 Sigmoid function allows each class to make a class-exclusion decision independently, 
while with SoftMax function, the probability of a class inevitably affects the probabilities of the remaining  classes. 
The criterion for excluding irrelevant classes based on the output of the Sigmoid activation function will be explained later.



\subsection{Class-Exclusion Aware Training}

To adjust the backbone neural network to exclude as many irrelevant classes as possible at intermediate layers, we introduce a class-exclusion aware training technique. Specifically, when training the backbone network, the accuracy of each class-exclusion network is also considered. Accordingly, we modify the original cost function for training the backbone network as follows 
\begin{equation}
\label{eq:cost function}
  L =L_{CE}+ \sum_{i=1}^{N}\sum_{j=1}^{M}L_{ij} 
\end{equation}
where $L_{CE}$ is the original cross entropy cost function. $N$ is the number of exit points in the backbone neural network. $M$ is the number of classes for classification at each early-exit point. For example, assume that a backbone network has 5 convolutional layers to classify 10 classes and each convolutional layer has an early-exit point. In this case, $N$ is 5 and $M$ is 10. 

$L_{ij}$ in Equation~\ref{eq:cost function} is the cost function for the $j$th class-exclusion network in the $i$th early-exit point. 
Since the Sigmoid function is used as the activation function in each exit point to avoid  class interference, binary cross-entropy is used as the cost function as follows
\begin{equation}
\label{eq:costfunction1}
    L_{ij}= -\alpha \frac{1}{N-i+1} 
     (
     \hat{y}_{ij}\log(y_{ij}) + (1-\hat{y}_{ij})\log(1-y_{ij}))
\end{equation}
where $N$ is the number of exit points, same as that in Equation~\ref{eq:cost function}. $\hat{y}_{ij}$ and $y_{ij}$ are the true label and the output of the class-exclusion network, respectively. 
$\alpha$ is a scaling factor used to prevent the vanishing gradient problem. 
Different exit points have different coefficients
$\frac{1}{N-i+1}$, with the coefficients of earlier layers are smaller than those in later layers. Such a different coefficient setting is used to alleviate the gradient imbalance issue \cite{li2019improved}. 


\subsection{Class-Exclusion Strategy for Dynamic Inference}

\begin{figure}
\centering    
\includegraphics{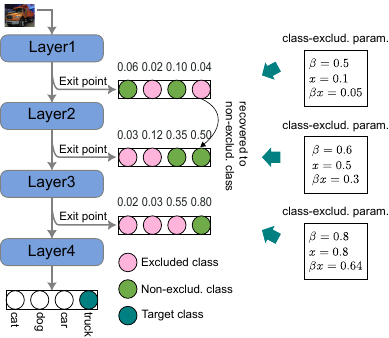}
\caption{
Class exclusion based on a dynamic threshold. } 
\label{fig:take_back}
\end{figure}  

To exclude irrelevant classes, we exploit the relative magnitude of probabilities generated by class-exclusion networks. 
The exploration of relative probability magnitudes realizes 
a dynamic threshold for class exclusion. 
Specifically, the maximum probability of all classes generated by class-exclusion networks, denoted as $x$, is identified at an intermediate layer. Afterwards, a class-exclusion coefficient, denoted as $\beta$, is determined by a search algorithm as described in the next subsection, is used to rule out irrelevant classes. Any class whose  probability output generated by its class-exclusion network is smaller than $\beta x$ will be excluded. 

To avoid that the target class is excluded unexpectedly at an intermediate layer, the class with the maximum probability generated by the class-exclusion network in the next layer is  identified. If this class is excluded in the previous layer, it is recovered and put back to the remaining classes in the next layer for further class exclusion. Figure \ref{fig:take_back} illustrates an example of class exclusion based on a dynamic threshold, where the truck class is excluded in the first layer and put back to the remaining classes in the second layer. 





To determine an optimal class-exclusion coefficient $\beta$ for each exit point, we develop a search algorithm to balance computational cost and inference accuracy. The search starts by initializing the class-exclusion coefficients of all exit points to 0. Afterwards, the number of MAC operations in each layer is ranked and the class-exclusion coefficient of the layer with the largest number of MAC operations is first searched since such a layer tends to have more computing resource to achieve a better class-exclusion capability than the other layers.  After the optimal $\beta$ for the exit point of this layer is determined, $\beta$ is fixed for this layer and the search of the next layer starts. The search process ends until $\beta$ in the layer with the smallest number of MAC operations is determined.


To search an optimal $\beta$ for an exit point, $\beta$ is increased by 0.01 from 0 and the accuracy of the class-based early-exit network is evaluated with this value in each iteration. When the accuracy degradation is larger than a specified value, the previous $\beta$ value is set as the class-exclusion coefficient for this exit point. 


%

\section{Experimental Results}\label{sec:results}
To demonstrate the effectiveness of the proposed class-aware early-exit framework in reducing the computational cost, three neural networks, 
AlexNet \cite{alexnet}, VGGsmall \cite{VGGsmall} and 
ResNet50 \cite{resnet} were trained and evaluated on CIFAR10 \cite{cifar} and CIFAR100 \cite{cifar}, respectively. 
For all neural networks, each convolutional layer has one exit point. For ResNet50, each layer consists of multiple residual blocks. 
The coefficient $\alpha$ in the class-exclusion aware cost function for early-exit was set to 36 for neural networks on CIFAR10 and 396 on CIFAR100.  
For the initial neural networks, AlexNet was trained by transferring the knowledge of a pre-trained AlexNet on ImageNet \cite{IM1k} into this model. During this process, a learning rate of 0.001 was set initially and the training epochs were set to 50. 
For VGGsmall and ResNet50, they were trained from scratch with an initial learning rate 0.001 and 200 epochs were used to train them.  
All experiments were conducted on an NVIDIA A100 80GB GPU.


    \begin{table}
    \footnotesize
    \centering
    \caption{Overall performance of the proposed framework}
    \vspace{-2pt}
    \begin{tabular}{ccccccc}
        \hline
         \multirow{2}{*}{NN-Dataset} &  \multicolumn{2}{c}{Acc. comp.}&  \multicolumn{3}{c}{ FLOPs (G)}\\
         \cmidrule(lr){2-3} \cmidrule(lr){4-6}
          & Ori. & Pro. & Ori. & Pro. & Red. \\
         \hline
         AlexNet-CIFAR10  & 90.54\% & 89.34\% & 1.4386 & 1.0375 & 27.88\% \\
         VGGs-CIFAR10 & 93.89\% & 91.93\%  & 1.5460 & 1.0668 & 31.00\% \\
         VGGs-CIFAR100 & 72.19\% & 71.11\% & 1.5463 & 1.1535 & 25.40\% \\
         ResNet50-CIFAR100 & 76.46\% & 74.39\% & 2.6008 & 1.7411 & 33.06\% \\
         \hline
    \end{tabular}
    \label{tab:drop1}
\end{table}

    Table \ref{tab:drop1} demonstrates the performance of the proposed early-exit framework. The used neural networks and their corresponding datasets are shown in the first column. The second and the third columns are the inference accuracy without and with applying the proposed framework, respectively. According to these two columns, there is only a slight accuracy loss of neural networks with the proposed framework. To verify the reduction of computational cost in executing neural networks, we compared the average number of FLOPs required to perform inference for an input with and without the proposed method. The FLOPs were evaluated by adding the numbers of multiplication, accumulation and operations incurred by average pooling. The results are shown in the last three columns. According to these results, the average computational cost in inference can be reduced significantly.

\begin{figure}
    \centering
    \includegraphics{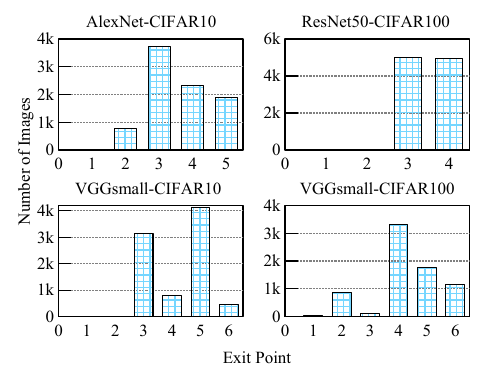}
    \caption{The number of input images that can be classified in the intermediate exit point of neural networks.} 
    \label{fig:amount of exited images}
    \end{figure}

   To demonstrate the effectiveness of the proposed early-exit framework, the number of images that can be classified in each exit point is evaluated. The results are shown in 
   Figure \ref{fig:amount of exited images}, where  
   the total number of test images is 10000 in each combination of neural networks and datasets. The x-axis shows the layer index. It can be clearly seen that, inputs tend not to be classified at the early layers of a neural network since the low-level features at such layers are not sufficient to achieve accurate classifications. For example, for AlexNet-CIFAR10 and VGGsmall-CIFAR10, there are no inputs that can be classified after the first layer. More inputs tend to be classified in the middle layers and the later layers since such layers can learn more complex features which are sufficient for accurate classification.
   
    \begin{figure}
    \centering    \includegraphics{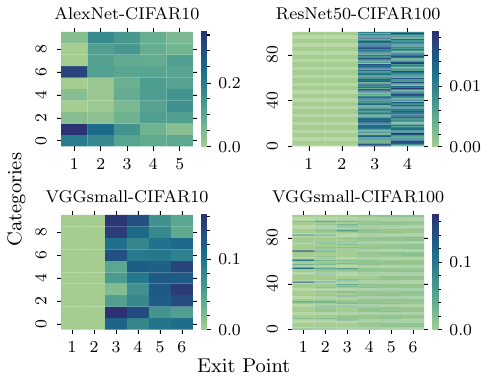}
    \caption[heatmap]{Percentage of classes that can exit early with respect to the exited images.  
    For CIFAR10, class 0-9 are airplane, automobile, bird, cat, deer, dog, frog, horse, ship and truck, respectively. }
    \label{fig:exitpercentage}
    \end{figure}   

  To show which type of classes can be classified in each exit point, we evaluated the percentage of each class that can exit early with respect to all exited images.
  The result is illustrated in 
   Figure \ref{fig:exitpercentage}, where x-axis is the layer index and y-axis is the class label, e.g., 0 corresponding to airplane. 
   Dark green indicates a large percentage while light green represents a small percentage. According to this figure, some classes tend to exit earlier than the other classes. 
  For example in VGGsmall-CIFAR10, most input images that can be classified at the 3rd layer are from class 1 (automobile) and class 9 (truck). In the exit points of the sixth layer, more classified images are from class 3 (cat) and class 5 (dog). This difference provides some hints that the difficulty of classifying images of cat and dog is higher than that of classifying images of automobile and truck.


    \begin{figure}
    \centering
    \includegraphics{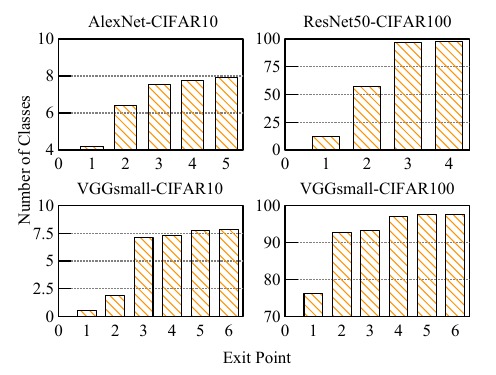}
    \caption{The average number of excluded classes in each exit point of intermediate layers in neural networks. }
    \label{fig:Exclude amount}
    \end{figure}
    \begin{figure}
    \centering
    \includegraphics{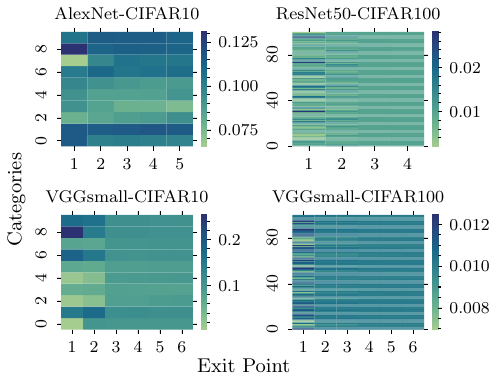}
    \caption{
    Percentage of classes that are excluded in each intermediate layer.  For CIFAR10, class 0-9 are airplane, automobile, bird, cat, deer, dog, frog, horse, ship and truck, respectively.
    }
    \label{fig:Exclude heat}
    \end{figure}

    To verify that the proposed method can exclude irrelevant classes at intermediate layers, we evaluated the average number of excluded classes at each intermediate layer. The results are illustrated in 
    Figure \ref{fig:Exclude amount}, where the number of excluded classes at a later exit point contains the number of excluded classes in the previous exit points. 
    According to this figure, it is clearly seen that 
   although a neural network can not make a decision at the shallow layers, it can still exclude some irrelevant classes. For example, in VGGsmall-CIFAR100, at the end of the first exit point, 75 classes can already be excluded dynamically. It can also be observed that the ability of neural networks to exclude classes is improved significantly after going through a few layers.  For example in AlexNet-CIFAR10, at the first exit point, the neural network can hardly exclude any classes. At the second exit point, it can exclude more than six classes. In addition, the ability of neural networks to exclude classes grows very slowly in later layers, indicating that a neural network has to spend more effort to classify similar classes. 


    Figure \ref{fig:Exclude heat} shows the percentage of each excluded class with respect to all excluded classes at each exit point, where 
    dark color represents a large percentage and light color represents a small percentage. 
    According to this figure, among the excluded classes, some classes are easier to be excluded than the other classes. For example, in VGGsmall-CIFAR10, most of the excluded classes in the first and second exit points come from class 8 (ship), class 6 (frog) and class 1 (automobile).
This observation also indicates that the features of these classes are significantly different from the other classes, so that they can be classified more easily at early layers. Figure \ref{fig:exitpercentage} also confirms this, where the images of these three classes tend to be classified at the exit point 3. 
   
   

    \begin{figure}
    \centering
    \includegraphics{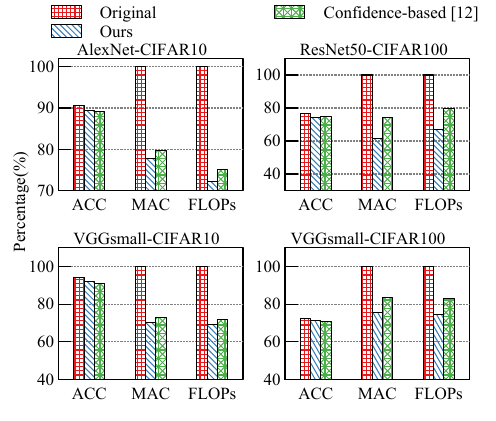}
    \caption{Comparison in accuracy, the remaining number of MAC operations and FLOPs with the previous method \cite{branchynet}.}
    \label{fig:compare}
    \end{figure}

   To demonstrate the advantages of the proposed method over the previous confidence-based early-exit strategy, we compared the accuracy, the remaining number of MAC operations and FLOPs between the proposed method and the previous method \cite{branchynet}. To fairly compare the performance, we adapted the method in \cite{branchynet} by adding early-exit in each layer and do not revise the original network structures. 
   The relative comparison results are shown in Figure \ref{fig:compare}, where red bar, blue bar and green bar represent the original neural network,  the proposed method and the confidence-based early exit, respectively. 
   Under almost the same accuracy, the remaining number of MAC operations and FLOPs with the proposed method is smaller than that from the previous confidence-based early-exit method.


\section{Conclusion}
\label{sec:conclusion}
In this paper, we have proposed a class-based early-exit to realize dynamic inference for DNNs to reduce the computational cost. Specifically, we take advantage of the learned features at intermediate layers to exclude as many irrelevant classes as possible, %
so that later layers only have to determine the final class among the remaining classes. 
Until there is only one remaining class at a layer, this class is the corresponding correct classification result. 
Experimental results demonstrate the FLOPs in executing DNNs in inference can be reduced by up to 33.06\%.
\label{sec:conclusion}

\let\oldbibliography\thebibliography
\renewcommand{\thebibliography}[1]{%
\oldbibliography{#1}%
\fontsize{7.0pt}{7.0}\selectfont
\setlength{\itemsep}{0.4pt}%
}

\bibliographystyle{IEEEtran}
\bibliography{IEEEabrv,CONFabrv,bibfile}

\end{document}